# A New Approach in Persian Handwritten Letters Recognition Using Error Correcting Output Coding

**Maziar Kazemi[✉1], Muhammad Yousefnezhad[2], Saber Nourian[3]**

1) Master Student at Department of Computer Science, University College of Rouzbehan, Sari, Iran
2) Research Assistant of Computer Science and Technology, Nanjing University of Aeronautics and Astronautics, Nanjing, China
3) Department of Electrical Engineering, Sari Branch, Islamic Azad University, Sari, Iran
maziar.kazemi@rouzbahan.ac.ir; myousefnezhad@nuaa.edu.cn; snourian@elec.iust.ac.ir

**Abstract**

*Classification Ensemble, which uses the weighed polling of outputs, is the art of combining a set of basic classifiers for generating high-performance, robust and more stable results. This study aims to improve the results of identifying the Persian handwritten letters using Error Correcting Output Coding (ECOC) ensemble method. Furthermore, the feature selection is used to reduce the costs of errors in our proposed method. ECOC is a method for decomposing a multi-way classification problem into many binary classification tasks; and then combining the results of the subtasks into a hypothesized solution to the original problem. Firstly, the image features are extracted by Principal Components Analysis (PCA). After that, ECOC is used for identification the Persian handwritten letters which it uses Support Vector Machine (SVM) as the base classifier. The empirical results of applying this ensemble method using 10 real-world data sets of Persian handwritten letters indicate that this method has better results in identifying the Persian handwritten letters than other ensemble methods and also single classifications. Moreover, by testing a number of different features, this paper found that we can reduce the additional cost in feature selection stage by using this method.*

***Keywords:*** *Persian Handwritten Letters Recognition, Classification Ensemble, Feature Selection, Accuracy, Support Vector Machine*

## 1. Introduction

Nowadays, the pattern recognition has the functional role in many contexts. Optical Character Recognition (OCR) is one of the branches of pattern recognition, which is widely used in machine learning literature [1]. Identifying handwritten letters is also a subset of OCR [2]. In the last two decades, there were a lot of extensive activities done to use a computer for reading the handwritten texts [3] and some of these methods such as template matching are sensitive to size and displacement and need to be normalized, so we are looking for algorithms that don't need to be normalized [4]. Many researches have been done in the field of OCR, which is basically for identifying the English numbers and letters, and good results have been achieved in this area. But in the case of Persian and Arabic due to their special features [5], the designed algorithms and software have less accuracy and less diagnostic power. A limited range of researches have been done in the context of identifying the Persian letters (especially handwritten letters) and there are some barriers.

Generally, different models provide predictions with different accuracy rates. Thus, it would be more efficient to develop a number of models using, different data subsets, or utilizing differing conditions within the modeling methodology of choice, to achieve better results. [6]. The general idea is shown in figure 1:

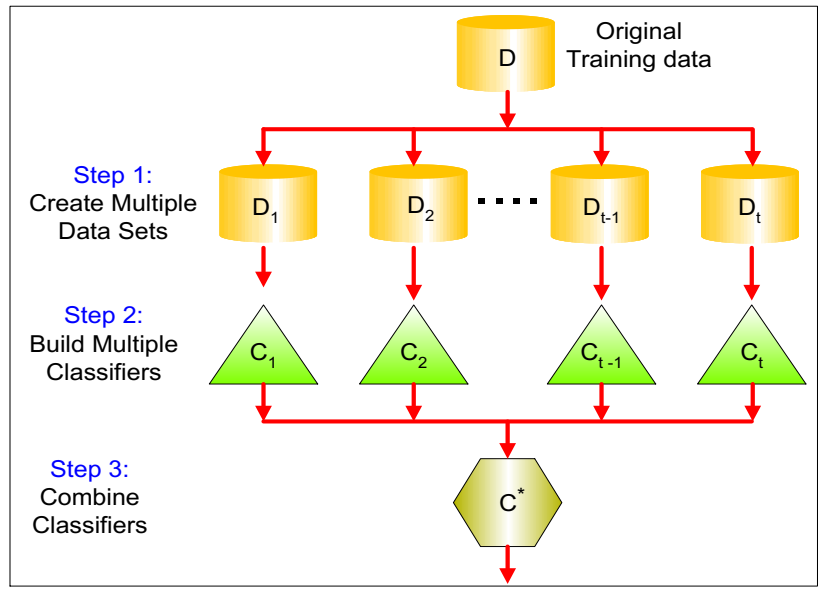

*Figure 1: the general idea of combining methods*

If we have some basic classifiers, we can achieve a higher accuracy by combining their results. The idea of ensemble methods is to create a set of classifiers using training data, and we obtain the level of accuracy by performing voting operations on results. Selecting the best model is not necessarily the ideal choice because potentially valuable information may be wasted by discarding the results of less-successful models. This leads to the concept of combining, where the outputs (individual predictions) of several models are pooled to make a better decision (collective prediction) [6]. In most cases, the combining methods have better results than single classifiers [7].

The aim of this paper is to use ECOC method in order to identify the Persian handwritten letters and to compare it with conventional methods of classification and combination. The error-correcting output coding (ECOC) method provides a more robust way for handling multiclass problems. The ECOC technique has demonstrated to be able to decrease the error caused by the bias and the variance of the base learning algorithm. The method is inspired by an information-theoretic approach for sending messages across noisy channels. The idea behind this approach is to add redundancy into the transmitted message by means of a codeword, so that the receiver may detect errors in the received message and perhaps recover the original message if the number of errors is small. ECOC is a successful method for solving the problems of multi-class learning using combined sets of binary classifications [8].

The usual way to proceed is to reduce the complexity of the problem by dividing it into a set of multiple simpler binary classification sub problems. One-versus-one (pairwise) or one-versus-all grouping voting techniques or trees of nested dichotomies [9] are some of the most frequently used schemes. In the line of the aforementioned techniques Error Correcting Output Codes [10] were born. ECOC represents a general framework based on a coding and decoding (ensemble strategy) technique to handle multiclass problems. One of the most well-known properties of the ECOC is that it improves the generalization performance of the base classifiers [11]. Moreover, the ECOC technique has demonstrated to be able to decrease the error caused by the bias and the variance of the base learning algorithm.

In the ECOC technique, the multiclass to binary division is handled by a coding matrix. Each row of the coding matrix represents a codeword assigned to each class. On the other hand, each column of the matrix (each bit of the codeword) shows a partition of the classes in two sets. The ECOC strategy is divided in two parts: the coding part, where the binary problems to be solved have to be designed, and the decoding technique, that given a test sample looks for the most similar codewords. Very few attention has been paid in the literature to the coding part of the ECOC. The most known coding strategies are one-versus-all, all-pairs (one-versus-one) and random coding.

Crammer et. al [12] were the first authors reporting improvement in the design of the ECOC codes. However, the results were rather pessimistic since they proved that the problem of finding the optimal discrete codes is computationally unfeasible since it is NP-complete. Specifically, they proposed a method to heuristically find the optimal coding matrix by changing its representation from discrete to continuous values. Recently, new improvements in the problem-dependent coding techniques have been presented by Pujol et al. [13]. They propose embedding of discriminant tree structures in the ECOC framework showing high accuracy with a very small number of binary classifiers, still the maximal number of dichotomies is bounded by the classes to be analyzed.

Main advantages of Ensemble learning methods are:
1- **Reduced variance**:  results are less dependent on peculiarities of a single learner and training set.
2- **Reduced bias**: combination of multiple classifiers may produce more reliable classification than single classifier.

Our contribution in this paper can be summarized as follows:
1- ECOC is easy to build and faster to predict.
2- Resistance to over training and over-fitting of data.
3- Ability to handle data without preprocessing or rescaling.
4- Resistant to outliers and can handle missing values.

The rest of this paper is organized as follows: Section 2 reviews some relevant literature. In Section 3, the basic classification methods are explained, and the classical ensemble methods are explained in Section 4 Then in section 5, the proposed model is presented and in section 6, the assessment and review of benefits of the proposed model will be discussed and finally in section 7, the results of this study and the policy of future work will be expressed.

## 2. Related Works

In this section we will discuss about an overview of works done in the field of identifying the handwritten letters of Persian and Arabic.

Masruri [14] has provided a method for recognition of separate Persian handwritten letters. He has considered 33 classes for Persian letters. Recognition of letters will be done in two stages. In the first stage, the letters get divided into eight groups using a "principal-basis" fuzzy classifier which its rules are learned by a set of educational examples. The features which are used at this stage are obtained by landmark approach in the directions of 45 and 90 degrees of binary image. In the second stage, the final recognition of letters in each group gets done through decision tree classifiers that are specifically designed for each group. At this stage, applied features are generally structural.

Salehian et al. in [15] have provided a complete system for identifying "Nastaliq" Persian words using neural networks. At the pre-processing stage, after finding the connected parts, they discovered the strokes of letters and removed them from image and by using a scanning algorithm which works based on upper and lower contour of the word; they divided the word's image to a sequence of sub-words. After the scanning operation, eight characteristics containing three descriptors of Fourier and a number of structural features for displaying sub-words were used in featured space. Identification was performed using a perceptron neural network.

Molaii et al. [16] have suggested a method for recognizing Iran's postal codes and city names which are handwritten on postal packages. The feature vectors of this method are calculated by using Discrete Wavelet Transform (DWT) with Haar Wavelet basis. In the feature-extraction process, three-level wavelet transform is applied on the thin image. A MLP neural network with back-propagation error rule is used for teaching the system.

Ebranhinpur et al. [17] have provided a method based on Stack Generalization method that named Modified Stack Generalization. They experiments have been done on 780 samples of 30 city names of Iran that for different experiments different number of training and testing samples was chosen. In the feature extraction Stage Gradient, Zoning methods are used, and also other method base on Gradient is suggested. Results show that Modified Stack generalization method with the recommended feature extraction method has been achieved to 92.21% recognition rate.

Borna et al. [18] have suggested aims to improve the feature extraction of Persian handwritten number recognition systems. They introduced nine new features for detection and recognition of Persian handwritten digits using the technique of finding the smallest enclosing disc in computational geometry. All these features are based on the geometry form of numbers and are much better than the features in terms of accuracy such as gradient which are mentioned as the strongest feature in the literature. Due to the improvement of recognition rate and acceptable speed.

## 3. Basic classifications

Selecting the type of classification model is one of the stages of pattern recognition. Classifier is considered as the core of a pattern recognition system [19]. The classifier attributes each unknown pattern to one of the known classes based on its characteristics. After selecting the model, the parameters must be specified. Parameters are determined during the learning process. Once the model is complete, using the test samples, pattern recognition system can be validated. In this part, we will explain about SVM classifier and other single classifiers.

*3.1 Support Vector Machine (SVM)*

SVMs works by finding a boundary in the feature space which maximizes the distance between feature vectors belonging to two distinct input classes [20]. The decision boundary usually takes the form of linear function which separates the two classes. In linearly inseparable problems, a non-linear decision surface is created by lifting the feature space into a higher dimensional space which allows a linear separating hyper plane to find [21]. This hyper plane corresponds to a non-linear decision surface in the original feature space. The mapping is denoted by $\varphi(x)$ which represents the map to the higher dimensional space where the data are linearly separable.

By using the kernel function $K(x, x_i) = \varphi(x_i)^T \varphi(x)$, the decision function of the SVM can be represented by:

$$f(x) = \sum_{i=1}^{n} \alpha_i y_i K(x, x_i) + b \tag{1}$$

Where $f(x)$-is the decision output, $y_i$ is the labelof the training symbol $x_i$ and $x$ is the symbols to be classified.

The parameters $\alpha_i$ and $b$ are found during training which is performed by solving the following optimization problem:

$$\min_{w,b} \frac{1}{2}||w||^2 + C \sum_{i=1}^{n} \epsilon_i \ subject \ to \ y_i(W^T \varphi(x_i) + b) \geq 1 - \epsilon_i \tag{2}$$

Many kernel functions exist but a well-performing kernel, used in many optical character recognition (OCR) systems, is the radial basis function (RBF):

$$K(x, x_i) = \exp\left(-\gamma ||x - x_i||^2\right), \gamma > 0 \tag{3}$$

The constant $C$ in Equation 2 is the penalty parameter of the error term and the constant $\gamma$ in Equation 3 is a kernel parameter. Both of these parameters have a significant effect on the accuracy of the trained system and need to be carefully set prior to the training process.

Although SVMs are binary classifiers, multi-class classification is easily achieved by combining SVMs in a one-against-others or one-against-one scheme [22]. Although SVM training time is proportional to the square of the number of samples and thus relatively slow, actual classification is very fast and can be performed in real-time [22].

*3.2 k-Nearest Neighbor*

K-nearest neighbor algorithm (k -NN) is a method for classifying objects based on closest training examples in the feature space. k -NN is a type of instance-based learning, or lazy learning where the function is only approximated locally, and all computations are deferred until classification. The k -nearest neighbor algorithm is amongst the simplest of all machine learning algorithms: an object is classified by a majority vote of its neighbors, with the object being assigned to the class most common amongst its k nearest neighbors (k is a positive integer, typically small). If k = 1, then the object is simply assigned to the class of its nearest neighbor.

*3.3 Decision Tree Learning*

DT as a machine learning tool uses a tree -like graph or model to operate deciding on a specific goal. DT learning is a data mining technique which creates a model to predict the value of the goal or class based on input variables. Interior nodes are the representative of the input variables, and the leaves are the representative of the target value. By splitting the source set into subsets based on their values, DT can be learned. Learning process is done for each subset by recursive partitioning. This process continues until all remain features in the subset has the same value for our goal or until there is no improvement in Entropy. Entropy is a measure of the uncertainty associated with a random variable.

*3.4 Artificial Neural Network*

An Artificial Neural Network (ANN) is a model which is to be configured to be able to produce the desired set of outputs, given an arbitrary set of inputs. An ANN generally composed of two basic elements: (a) neurons and (b) connections. Indeed, each ANN is a set of neurons with some connections between them. From another perspective, an ANN contains two distinct views: (a) topology and (b) learning. The topology of an ANN is about the existence or nonexistence of a connection. The learning in an ANN is to determine the strengths of the topology connections. One of the most representatives of ANNs is Multilayer Perceptron.

## 4. Ensemble Methods

An ensemble of classifiers is a collection of several classifiers whose individual decisions are combined in some way to classify the test examples. It is known that an ensemble often shows much better performance than the individual classifiers that make it up. Hansen et al. [17] shows why the ensemble shows better performance than individual classifiers as follows. Assume that there is an ensemble of n classifiers: $\{f_1, f_2, \ldots, f_n\}$ and consider a test data x. If all the classifiers are identical, they are wrong at the same data, where an ensemble will show the same performance as individual classifiers. However, if classifiers are different and their errors are uncorrelated, then when $f_i(x)$ is wrong, most of the other classifiers except for $f_i(x)$ may be correct. Then, the result of majority voting can be correct. More precisely, if the error of individual classifier is $p < 1/2$ and the errors are independent, then the probability $p$ that the result of majority voting is incorrect is:

$$\sum_{k=\left[\frac{n}{2}\right]}^{n} p^k (1-p)^{(n-k)} \left( < \sum_{k=\left[\frac{n}{2}\right]}^{n} \left(\frac{1}{2}\right)^k \left(\frac{1}{2}\right)^{(n-k)} = \sum_{k=\left[\frac{n}{2}\right]}^{n} \left(\frac{1}{2}\right)^n \right) \qquad (4)$$

When the size of classifiers $n$ is large, the probability $p$ becomes very small. The SVM has been known to show a good generalization performance and is easy to learn exact parameters for the global optimum. Because of these advantages, their ensemble may not be considered as a method for improving the classification performance greatly. However, since the practical SVM has been implemented using the approximated algorithms in order to reduce the computation complexity of time and space, a single SVM may not learn exact parameters for the global optimum. Sometimes, the support vectors obtained from the learning is not sufficient to classify all unknown test examples completely. So, we cannot guarantee that a single SVM always provides the global optimal classification performance over all test examples.

To overcome this limitation, we propose to use an ensemble of support vector machines. Similar arguments mentioned above about the general ensemble of classifiers can also be applied to the ensemble of support vector machines. Figure 2 shows a general architecture of the proposed SVM ensemble.

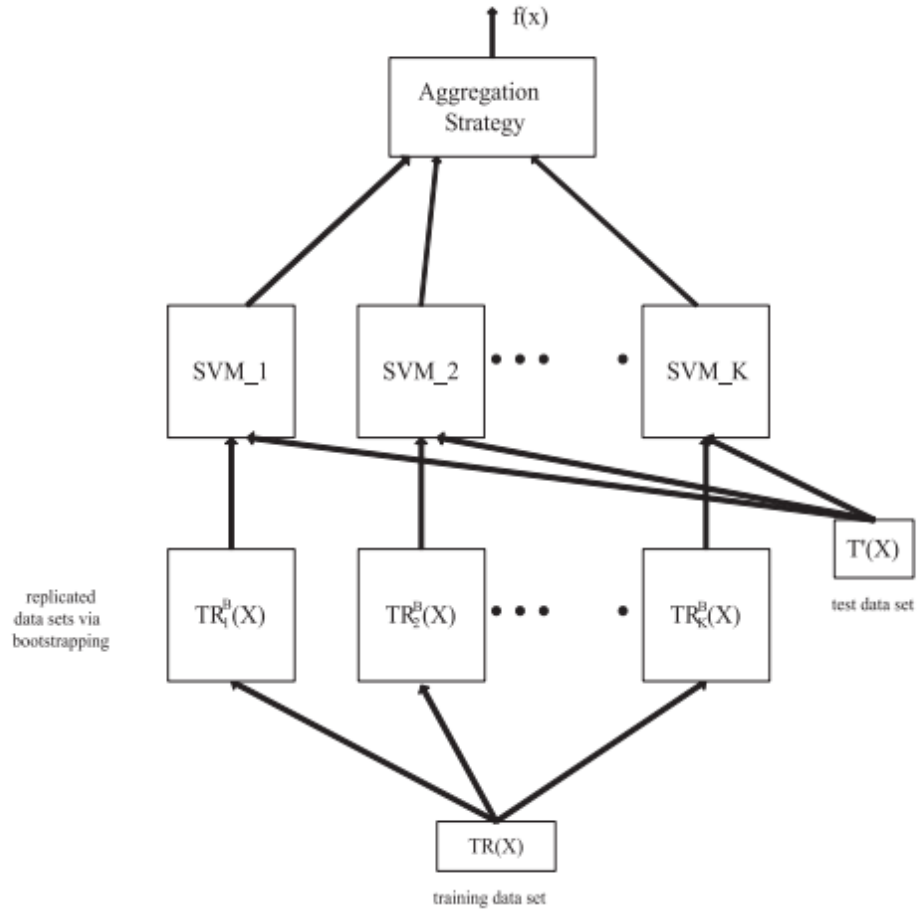

*Figure 2: A general architecture of the SVM ensemble*

### 4.1 Bagging

First, we explain a bagging technique to construct the SVM ensemble. In bagging, several SVMs are trained independently via a bootstrap method, and then they are aggregated via an appropriate combination technique. Usually, we have a single training set $TR = \{(x_i, y_i) | i = 1, 2, \ldots, l\}$. But we need K training samples sets to construct the SVM ensemble with K independent SVMs. From the statistical fact, we need to make the training sample sets different as much as possible in order to obtain higher improvement of the aggregation result. For doing this, we often use the bootstrap technique as follows.

Bootstrapping builds K replicate training data sets $\{TR_{bootstrap_k} | k = 1, 2, \ldots, K\}$ by randomly re-sampling, but with replacement, from the given training data set TR repeatedly. Each example $x_i$ in the given training set TR may appear repeated times or not at all in any particular replicate training data set. Each replicate training set will be used to train a certain SVM.

*4.2 Boosting*

The representative boosting algorithm is the AdaBoost algorithm. Like bagging, each SVM is also trained using a different training set. But the selection scheme of training samples in the AdaBoost method is quite different from the bagging method by the following. Initially, we have a training set $TR = \{(x_i, y_i) | i = 1, 2, \ldots, l\}$ consisting of $l$ whole samples, and each sample in the $TR$ is assigned to have the same value of weight $p_0(x_i) = 1/l$. For training the $k$th SVM classifier, we build a set of training samples $TR_{boost_k} = \{(x_i, y_i) | i = 1, 2, \ldots, l'\}$ that is obtained by selecting $l'(< l)$ samples among the whole data set $TR$ according to the weight values $p_{k-1}(x_i)$ at the $(k-1)$th iteration. These training samples are used for training the $k$th SVM classifier. Then, we evaluate the classification performance of the $k$th trained SVM classifier using the whole training sample $TR$ as follows. We obtain the updated weight values $p_k(x_i)$ of the training samples in $TR$ based on the errorless of the training samples as follows. The weight values of the incorrectly classified samples are increased, but the weight values of the correctly classified samples are decreased. This implies that the samples which are hard to classify are selected more frequently. This updated weight values will be used for building the training samples $TR_{boost_{k+1}} = \{(x_i, y_i) | i = 1, 2, \ldots, l'\}$ of the $(k+1)$th SVM classifier. This sampling procedure will be repeated until $K$ training samples set has been built for the $K$th SVM classifier.

## 5. Proposed model

Error-correcting output coding is a recipe for solving multi-way classification problems. It works in two stages: first, independently construct many subordinate classifiers, each responsible for removing some uncertainty about the correct class of the input; second, apply a voting scheme to decide upon the correct class, given the output of each weak learner. The next section introduces the technique of error-correcting output coding.

For multiclass learning, each class is represented by a unique bit string of length $n$ known as its codeword. We then train $n$ binary classifiers to predict each bit of the codeword string. The predicted class of a test instance is given by the codeword whose hamming distance between a pair of bit strings is given by the number of bits that differ. The flowchart of ECOC method, as illustrated in figure 3:

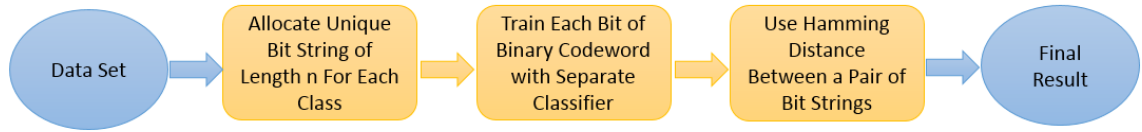

*Figure 3: Flowchart of ECOC method*

The ECOC designs are independent of the base classifier applied. They involve error-correcting properties and have shown to be able to reduce the bias and variance produced by the learning algorithm. Because of these reasons, ECOCs have been widely used to deal with multi-class categorization problems.

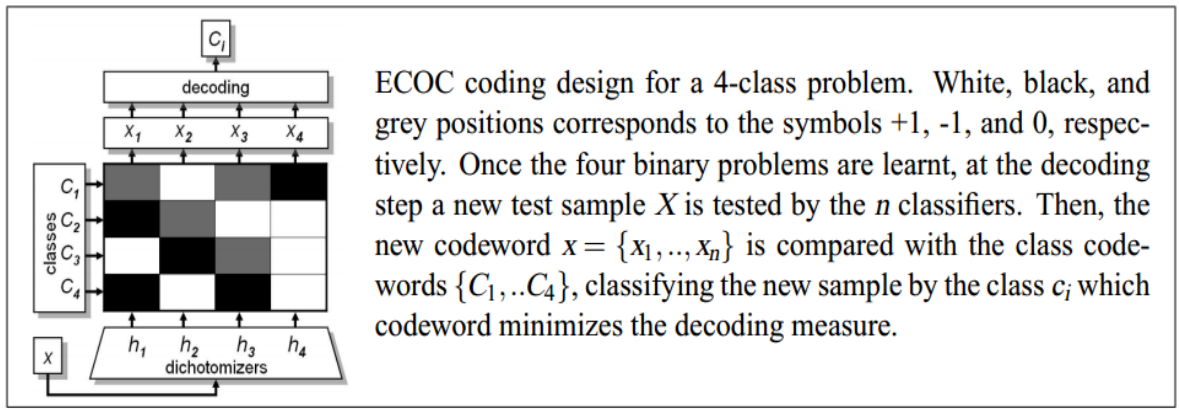

ECOC coding design for a 4-class problem. White, black, and grey positions corresponds to the symbols +1, -1, and 0, respectively. Once the four binary problems are learnt, at the decoding step a new test sample $X$ is tested by the $n$ classifiers. Then, the new codeword $x = \{x_1, .., x_n\}$ is compared with the class codewords $\{C_1, ..C_4\}$, classifying the new sample by the class $c_i$ which codeword minimizes the decoding measure.

*Figure 4: ECOC design example*

### 5.1 Error Correcting Output Coding (ECOC)

We describe the technique of error-correcting output coding with a simple example: the task of classifying news wire articles into the m = 4 categories {**politics**, **sports**, **business**, **art**}. To begin, one assigns a unique n-bit vector to each label (where $n > log_2^m$):

| label | coding |
|---|---|
| politics | 0110110001 |
| sports | 0001111100 |
| business | 1010101101 |
| arts | 1000011010 |

One can view the $i$th bitvector as a unique coding for label $i$. For this reason (and others, which will soon become apparent), we'll refer to the set of bit vectors as a code and denote it by $C$. The $i$th row of $C$ we will write as $C_i$ and the value of $j$th bit in this row as $C_{ij}$.

The second step in constructing an ECOC classifier is to build an individual binary classifier for each column of the code—10 classifiers in all, in this case. The positive instances for classifier $j$ are documents with a label $i$ for which $C_{ij} = 1$. The third classifier, for instance, has the responsibility of distinguishing between documents whose label is **sports** or **arts** and those whose label is **politics** or **business**. Heeding to convention, we refer generically to any algorithm for predicting the value of a single bit as a "plug-in classifier" (PiC). A PiC, then, is a predictor of whether a document belongs to some fixed subset of the classes.

> *Input*:   Documents $\{x_1, x_2, \dots, x_D\}$;
>    Labelings $\{y_1, y_2, \dots, y_D\}$ (with $m$ distinct labels);
>    Desired code size $n \geq log_2^m$
> *Output*:  $m$ by $n$ coding matrix $C$;
>    $n$ classifiers $\{\lambda^1, \lambda^2 \dots \lambda^n\}$
> 1. Generate a $m$ by $n$ 0/1 coding matrix $C$
> 2. Do for $j \in [1,2 \dots n]$
>    - Construct two superclasses, $S_j$ and $S'_j$. $S_j$ consists of all labels $i$ for which $C_{ij} = 1$, and $S_j$ is

*Figure 5: Algorithm 1, Training an ECOC document classifier*

To summarize, training an ECOC classifier consists of learning a set $\Lambda(x) = \{\lambda^1, \lambda^2 \dots \lambda^n\}$ of independent binary classifiers. With $\Lambda$ in hand, one can hypothesize the correct class of an unlabeled document x as follows. Evaluate each independent classifier on x, generating a n-bit vector $\Lambda(x) = \{\lambda^1(x), \lambda^2(x), \dots, \lambda^n(x)\}$. Most likely, the generated bit vector $\Lambda(x)$ will not be a row of $C$, but it will certainly be closer (in Hamming distance $\Delta$, say) to some rows than to others. Categorizing the document $x$ involves selecting $argmin_i \Delta(C_i, \Lambda(x))$, The label $i$ for which $C_i$ is closest to $\Lambda(x)$. (If more than one row of $C$ is equidistant to $\Lambda(x)$, select one arbitrarily.) For instance, if the generated bit vector $\Lambda(x) = \{1010111101\}$, the document would receive the label **business**.

To the extent that rows of $C$ are well-spaced in Hamming distance, the classifier will be robust to a few errant PiCs. This is the idea behind error-correcting codes as well: to transmit a point in the m-dimensional cube reliably over a noisy channel, map it to one of a set of well separated "fixed points" in a higher-dimensional cube; to recover the original point, find the closest fixed point to the point actually received and take its pre-image in the original cube.

> *Input*:   Training ECOC classifier: $m$ by $n$ coding matrix $C$ and $n$ classifiers $\{\lambda^1, \lambda^2 \ldots \lambda^n\}$;
> Unlabeled document $x$
> *Output*:   Hypothesized label $y$ for $x$
> 1. Do for $j \in [1,2 \ldots n]$
>    - Compute $\lambda^j(x)$---the confidence with which PiC $j$ believes $x \in S_j$.
>    - Output $\arg\min_i \sum_{j=1}^{n} |\lambda_j(x) - C_{ij}|$, $i \in [1,2 \ldots n]$

*Figure 6: Algorithm 2, Applying an ECOC document classifier*

In general, $\lambda^j(x)$ may not be a 0/1 value, but a realvalued probability, measuring the classifiers confidence that document $x$ belongs in the $j$th superclass.

## 6. Evaluation

In this section the results of applying the proposed method on different data sets are reported. In pre-processing stage, an operation was performed on images so that their quality is increased. In the feature extraction stage, we used five different methods, which were used in paper for recognition of handwritten letters and numbers. These methods include: Principal Component Analysis (PCA) [24], gradient of image [25], Gaussian filters [26], DCT coefficients [27] and Torques and GLCM matrix [28]. After data training using aforementioned methods and comparing the obtained results, we found out that PCA is the best method for this cause, and we used it in the experimental evaluation.

### 6.1 Data sets

The proposed method has been tested on 10 standard datasets. Each set contains 3200 samples of Persian handwritten letters, and they include 100 samples for each Persian letter. From each data set, 2240 samples are related to training phase (70 samples for each letter), and 960 samples are related to testing phase (30 samples for each letter). Each image contains a white background which the letters are in the middle of it. First to improve the quality, we converted the images to binary ones. In order to reduce the volume of calculations, we separate the letters from this background, and then we turn them into a standard size. Some of these data are shown in figure 7:

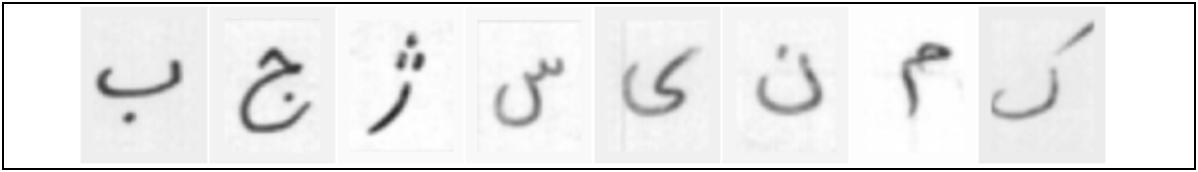

*Figure 7: Some examples of used data*

More information is available in [29].

*6.2 The experimental results*

The proposed method is implemented and tested in MATLAB (8.1) environment and the results of experimenting on average of 10 times of independent performance for each technique (Support Vector Machine (SVM), Decision Tree (DT), k-Nearest-Neighbor (KNN), Neural Network (NN), Bagging and Boosting) are reported. The results of this experiment are shown in table 1.

*Table 1: Test results ± standard error for 10 times of independent performance*

|  | SVM | DT | KNN | NN | Bagging | Boosting | ECOC |
|---|---|---|---|---|---|---|---|
| Random 1 | 77.01±0.415 | 52.27±0.639 | 72.92±0.481 | 54.06±0.633 | 72.71±0.434 | 81.57±0.331 | **88.04 ± 0.107** |
| Random 2 | 78.42±0.419 | 56.94±0.155 | 78.65±0.454 | 55.94±0.569 | 75.68±0.227 | 83.69±0.258 | **89.84±0.106** |
| Random 3 | 79±0.285 | 59.11±0.562 | 76.15±0.351 | 53.13±0.618 | 76.55±0.236 | 82.42±0.256 | **88.73±0.057** |
| Random 4 | 79.38±0.425 | 60.12±0.581 | 79.17±0.294 | 60.63±0.414 | 79.47±0.243 | 86.74±0.265 | **90.6±0.086** |
| Random 5 | 76.51±0.371 | 56.34±0.609 | 71.77±0.677 | 54.38±0.449 | 69.89±0.208 | 81.87±0.456 | **86.12±0.084** |
| Random 6 | 77.26±0.328 | 52.87±0.483 | 77.19±0.319 | 51.88±0.456 | 75.68±0.224 | 83.4±0.268 | **87.03±0.089** |
| Random 7 | 79.31±0.435 | 55.67±0.576 | 78.54±0.483 | 51.25±0.601 | 78.92±0.153 | 85.73±0.42 | **88.39±0.076** |
| Random 8 | 79.06±0.449 | 56.50±0.658 | 76.56±0.553 | 51.25±0.395 | 75.72±0.186 | 82.75±0.235 | **89.45±0.09** |
| Random 9 | 79.14±0.346 | 58.67±0.702 | 77.29±0.344 | 59.38±0.634 | 73.22±0.273 | 84.78±0.283 | **89.71±0.053** |
| Random 10 | 75.56±0.351 | 54.74±0.558 | 76.46±0.341 | 46.25±0.508 | 71.59±0.335 | 83.42±0.312 | **89.35±0.113** |

In this paper, we used Error Correcting Output Coding (ECOC) for combining in which the SVM is used as the basic classifier. In table (1), the best results are shown in bold. We have used two combining methods other than ECOC. Bagging method shows a good performance for unstable classifiers that have small perturbation. The classifiers in which their outputs are changed with a small difference in input, are called unstable. This method does not cause good changes in output because the SVM is stable. But Boosting method provides better results. Comparing to ECOC method, the disadvantage of these two methods is that their run time is too much. The ECOC method besides the lower run time, the ECOC method produces better results than other two methods in recognition of handwritten letters. In general, we can say that combining methods do well for weak classifiers, but the more classifiers get strong, the less will be the effect of method.

The results of this experiment and to compare it with other methods are shown in table 2.

*Table 2: Test experimental Result and Other Methods to Persian Handwritten Letters Recognition*

|  | **The proposed method** | **Masruri [14]** | **Salehian et al [15]** | **Molaii et al [16]** | **Ebranhinpur et al [17]** |
|---|---|---|---|---|---|
| **Accuracy** | 88.73 | 92 | 93 | 91.81 | 92.21 |

In [14] 4950 samples (150 samples of each letter) written by different writers with different levels of education have been used. 3300 samples from this set (100 samples of each letter) have been used for learning and 1650 samples for the test which according to the reported results 92% of these samples were correctly recognized. The drawback of this method in comparison to the proposed method is the high execution time of it. Considering high number of fuzzy laws, this running time is predictable. Meanwhile, the number of samples tested in the mentioned method is lower, and isn't considered as a part of standardized data. This results in the higher percentage of accuracy in this method.

In [15] 320 words with the "Nastaliq" font are used. In this method and the method applied in [16], Neural networks have been used for the training. Since neural networks are very time consuming for large data sets and don't return good results for Persian handwritten letters recognition, using it is not appropriate.

In [17] 780 samples from 30 different cities have been used. Accuracy level of 92.21% is obtained. Considering that in this method a combination of neural networks have been used and the fact that neural networks themselves are time consuming for classification therefor using them in hybrid methods is not appropriate with regard to the execution time. Besides, different methods for feature extraction have been tested on our standardized data set and the PCA method had the best result, so this method has been used. The Advantage of this method over our proposed method is the higher accuracy percentage, but considering the higher number of our test samples, this is trivial.

The average accuracy percentage of each technique is shown in figure 8.

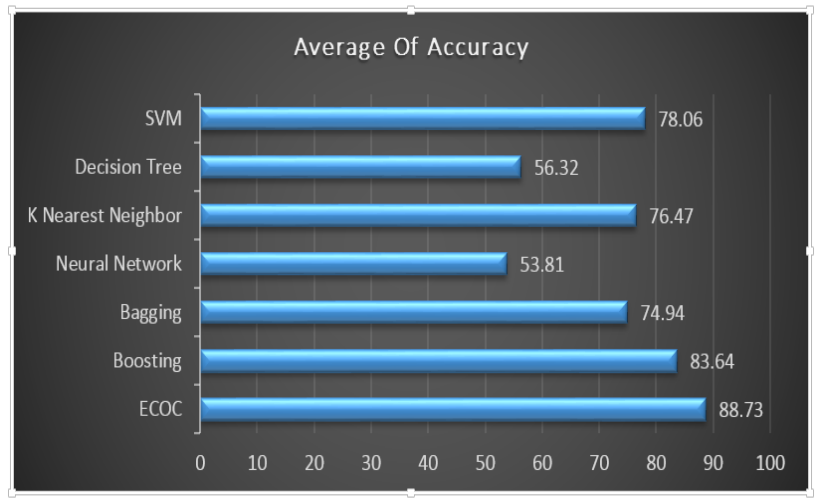

*Figure 8: Average of Accuracy for each technique in percentage*

According to this figure, the accuracy percentage of the proposed method (ECOC) is shown, which indicates the increase in average accuracy percentage in comparison to other methods. This paper presents an investigation of why the ECOC technique works, particularly when employed with SVM learning algorithms. It shows that the ECOC method (like any form of voting or committee) can reduce the variance of the learning algorithm. Furthermore (unlike methods that simply combine multiple runs of the same learning algorithm) ECOC can correct for errors caused by the bias of the learning algorithm.

One of the effective parameters in ECOC method is the length of code for each class. The accuracy levels of this method and different code lengths are given in figure 9.

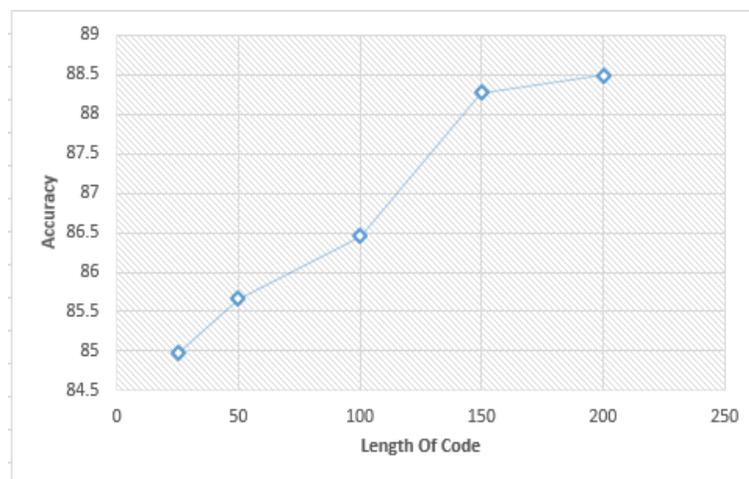

*Figure 9: accuracy percentage level for different code lengths*

According to this figure, results are getting better with an increase in code length till the point where changes remain almost constant. The reason for this, is that with an increase in code length, the generated code gets more unique. This power of distinguishing is getting higher. But it should be noted that increasing the code length raises program run time, and we should select a length of code, which would be proportional in both areas of accuracy percentage and run time. According to this figure, in this paper we have used the codes with the length of 150.

A simple calculation shows why the classifier performance improves with code length $n$. Assume for the moment that the PiCs only output binary values, and the errors committed by any two PiCs are independent of one another. Denote by $p_i$ the probability of error by the $i$th PiC, and let $p' \equiv max_i\ p_i$. If the minimum distance of $C$ is $\Delta_{min}$, then classification is robust to any $\lfloor \Delta_{min}/ 2 \rfloor$ or fewer errors, and so the probability of a correct classification, as a function of $n$, is

$$P(n) \geq \sum_{k=0}^{\lfloor \Delta_{min}/ 2 \rfloor} \binom{N}{k} {p'}^k (1-p')^{N-k} \qquad (5)$$

The quantity on the right—the first $\lfloor \Delta_{min}/ 2 \rfloor$ terms of the binomial expansion of $p + (1 - p)$—is monotonically increasing in $\Delta_{min}$, which itself increases with $n$ for a randomly-constructed code.

As it was mentioned before, PCA method is used for creating the feature vector in feature extraction stage. PCA is a statistical technique that selects a number of coefficients with the largest eigenvalue for each sample. The provided results, until now, are obtained from 20 coefficients of PCA. Figure 10 shows the average results of different techniques with the number of different features. According to figure 10, ECOC method provides better results for the number of different features.

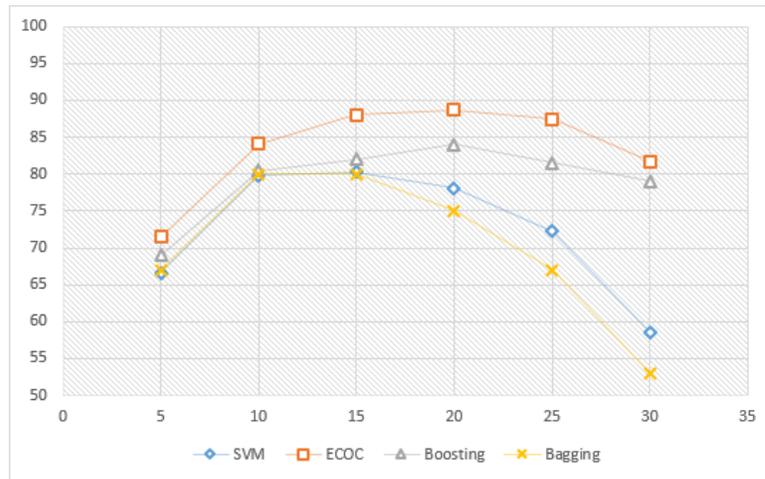

*Figure 10: average of obtained results for each technique using number of different features*

As we see in the picture, ECOC method provides similar results for numbers of 10 to 30, but we have different results for SVM. Therefore, one advantage of this method is that it can significantly reduce the time cost from feature selection stage.

## 7. Conclusion

In this paper, the ECOC ensemble method was used for identifying the Persian handwritten letters. Since there's no classification to make optimal results for every issue, we can achieve the desired response by combining them. The experimental results of this method for handwritten letters' recognition problem on 10 different data sets and comparing them with other methods such as neural networks, K the nearest neighbor, Boosting, Bagging and SVM shows that this method is superior in comparison with other conventional classification methods. The ECOC method provides better results, and also the running time is also suitable. Besides, this method provides almost the same and constant results for a number of different features. Two important achievement of this method which were examined in this paper: first, reaching to higher accuracy rate than other learning algorithms with good running time. Second, reducing the time of feature selection stage which these two features are considered as important factors for learning problems.

For future works, we suggest to use special codes (non-randomly) for each class to achieve more differences between classes and better results.